# Bench-RNR: Dataset for Benchmarking Repetitive and Non-repetitive Scanning LiDAR for Infrastructure-based Vehicle Localization[1]

Runxin Zhao, Chunxiang Wang, *Member, IEEE*,
Hanyang Zhuang, *Member, IEEE*, and Ming Yang, *Member, IEEE*

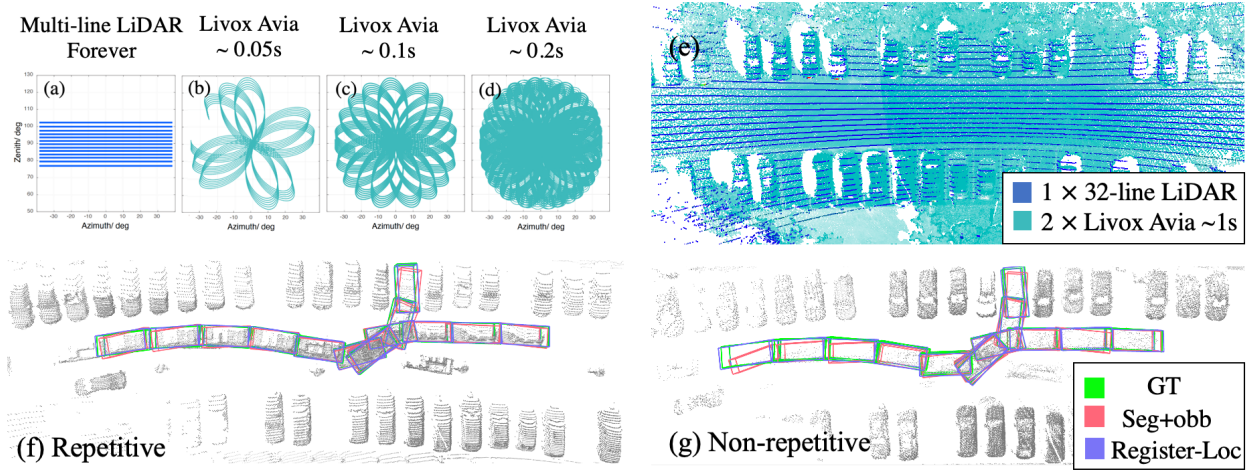

Fig. 1. Scanning pattern of rotating multi-line LiDAR (repetitive scanning LiDAR) and non-repetitive scanning LiDAR Livox Avia. (a) shows the point cloud of repetitive scanning LiDAR accumulated forever. (b)-(d) represent the point clouds of non-repetitive scanning LiDAR accumulated for 0.05s, 0.1s, and 0.2s, respectively. (e) shows the point clouds from a 32-line LiDAR and two Livox Avia for the same scene, with Avia's point cloud accumulated for 1s. (f) and (g) are vehicle localization results using infrastructure-based repetitive and non-repetitive scanning LiDAR respectively.

*Abstract*—Vehicle localization using roadside LiDARs can provide centimeter-level accuracy for cloud-controlled vehicles while simultaneously serving multiple vehicles, enhancing safety and efficiency. While most existing studies rely on repetitive scanning LiDARs, non-repetitive scanning LiDAR offers advantages such as eliminating blind zones and being more cost-effective. However, its application in roadside perception and localization remains limited. To address this, we present a dataset for infrastructure-based vehicle localization, with data collected from both repetitive and non-repetitive scanning LiDARs, in order to benchmark the performance of different LiDAR scanning patterns. The dataset contains 5,445 frames of point clouds across eight vehicle trajectory sequences, with diverse trajectory types. Our experiments establish baselines for infrastructure-based vehicle localization and compare the performance of these methods using both non-repetitive and repetitive scanning LiDARs. This work offers valuable insights for selecting the most suitable LiDAR scanning pattern for infrastructure-based vehicle localization. Our dataset is a significant contribution to the scientific community, supporting advancements in infrastructure-based perception and vehicle localization. The dataset and source code are publicly available at: https://github.com/sjtu-cyberc3/BenchRNR.

*Index Terms*— Dataset, Roadside LiDAR, Infrastructure-based Vehicle Localization, Intelligent Transportation Systems.

## I. INTRODUCTION

With the advancement of vehicle-infrastructure cooperation technology, an increasing number of scenarios now deploy roadside sensors to enhance the safety and efficiency of Intelligent Transportation Systems (ITS). Vehicle perception and localization based on roadside sensors can provide centi-

[1] This work was supported by National Natural Science Foundation of China (62203294/U22A20100/ 62373250). Hanyang Zhuang is the corresponding author.

Runxin Zhao, Chunxiang Wang, and Ming Yang are with Department of Automation, Shanghai Jiao Tong University, Shanghai, 200240; Key Laboratory of System Control and Information Processing, Ministry of Education of China, Shanghai, 200240; Shanghai Engineering Research Center of Intelligent Control and Management, Shanghai 200240, China (mingyang@sjtu.edu.cn). Hanyang Zhuang is with Global College, Shanghai Jiao Tong University, Shanghai, 200240, China (zhuanghany11@sjtu.edu.cn).

meter-level accuracy for cloud-controlled vehicles while simultaneously serving multiple vehicles, reducing the cost of installing expensive sensors on each individual vehicle.

Most current research employs rotating multi-line LiDAR (repetitive scanning LiDAR) to achieve infrastructure-based perception and vehicle localization. However, the vertical resolution of rotating multi-line LiDAR is limited by the number of its laser scanners, inevitably generating blind zones between scanning lines when installed in a fixed position (See Fig. 1.a). Alternatively, the novel non-repetitive scanning LiDAR, named for its unique scanning pattern, can progressively cover the entire field of view without repeating itself. This can eliminate fixed blind zones in roadside sensor configurations (see Fig. 1.b - d). Moreover, non-repetitive scanning LiDAR is more cost-effective than rotating multi-line LiDAR. A detailed comparison can be seen in Table 1.

TABLE 1. TYPICAL PROPERTIES OF REPETITIVE AND NON-REPETITIVE SCANNING LIDARS

| Item | Repetitive LiDAR | Non-repetitive LiDAR |
|---|---|---|
| Representative model | Hesai, Robosense | Livox |
| Point rate single return | 3,460k pts/s (Hesai OT128) 576k pts/s (RS Helios) | 240k pts/s (Livox Avia) 450k pts/s (Livox HAP) |
| Line number | 128 / 64 / 32 lines | Equivalent to 144 lines (Livox HAP) |
| Price | 140,000 $ (Hesai OT128) | 1,400 $ (Livox Avia) |
| Coverage | Fixed blind zones between lines | Entire FOV |

Some studies have demonstrated the potential advantages of non-repetitive scanning LiDAR [1], but it has not yet been widely used in roadside perception and localization tasks, nor has any open-source dataset containing roadside non-repetitive scanning data been released. To address this gap, our dataset, named Bench-RNR, collected data from both repetitive and non-repetitive scanning LiDARs for localization tasks. We also performed comparative experiments on the localization performance of these two LiDAR types, providing guidance for selecting LiDAR for roadside vehicle localization applications.

Bench-RNR includes LiDAR data from a 128-line repetitive scanning LiDAR and two non-repetitive scanning LiDARs, along with two cameras. It contains 5,445 frames, covering parking and driving trajectories across eight vehicle sequences with various trajectory types. High-precision temporal alignment and sensor calibration have been conducted. Additionally, we provide four baselines for infrastructure-based vehicle localization and a comparison of the localization results for non-repetitive and repetitive scanning LiDARs. Finally, our dataset has been open-sourced to facilitate use by the scientific community.

In summary, our contributions are:
(1) We present Bench-RNR, a diverse infrastructure-based LiDAR vehicle localization dataset with both repetitive and non-repetitive scanning LiDAR, which has undergone precise temporal synchronization and sensor calibration.
(2) We provide several typical baselines for infrastructure-based vehicle localization and evaluate their performance using Bench-RNR.
(3) We conduct comparative experiments on the localization performance of repetitive and non-repetitive scanning LiDAR. The results can provide guidance for the selection of LiDAR in infrastructure-based vehicle localization.

## II. RELATED WORKS

### A. Infrastructure-based dataset

With the advancement of V2X technology, infrastructure-side perception datasets have emerged. DAIR-V2X-I [2] is dedicated to roadside perception, containing 10,084 frames of jointly annotated images and 300-line LiDAR data collected at 28 intersections. The TUMTraf Intersection Dataset [3] includes 4.8k images and point clouds with over 57.4k manually labeled 3D boxes for Urban 3D Camera-LiDAR roadside perception. V2X-Real [4] releases a large-scale, multi-modal dataset from smart infrastructures. However, cloud-controlled vehicle localization requires centimeter-level accuracy to ensure safe operation, and these roadside datasets are primarily designed for 3D object detection, whose accuracy falls short of the standards needed for localization validation. Consequently, our dataset uses high-precision vehicle-end GNSS and IMU to establish ground truth, providing high-accuracy annotations for localization.

### B. Non-repetitive Scanning

Non-repetitive scanning is a LiDAR scanning pattern where the emitted laser does not trace the same trajectory during continuous scanning [5]. This enables denser point clouds through temporal integration, which is important for tasks like high-precision static scene modeling and environmental feature extraction [6], [7]. By acquiring point cloud data from both repetitive and non-repetitive LiDAR in simulations, [1] demonstrates the potential of non-repetitive scanning LiDAR for 3D object detection. Loam-livox [8] and Point-lio [9] use non-repetitive scanning LiDAR to simultaneously localize the robot's pose and build high-resolution maps of the surrounding environment. LidPose [10] utilizes it to detect moving pedestrians and fit 3D human skeleton, while M-detector [11] use it to achieve instantaneous detection of moving events.

In non-repetitive scanning LiDAR datasets, Livox has released an open-source dataset [12] focused on vehicle-side object detection. A. Xie et al. [1] also provided a dataset collected in the Carla simulation environment. However, there remains a domain gap between simulation and the real world. Currently, there is a lack of real-world non-repetitive scanning LiDAR datasets for infrastructure-side applications, as well as comparative studies between repetitive and non-repetitive LiDAR in this context. Our dataset is the first open-source infrastructure-side LiDAR vehicle localization dataset containing both repetitive and non-repetitive LiDAR data.

## III. BENCH-RNR DATASET

In this section, we introduce our methods for data collection, sensor calibration and data annotation, and then provide a brief analysis of the dataset.

### A. Data Collection

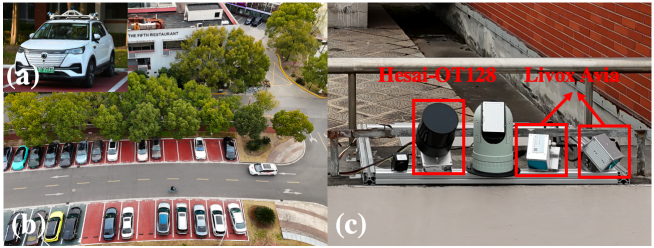

Fig. 2. (a): Experimental vehicle. (b) Bird-eye-view picture of the experimental site. (c). Sensor Suite.

We collected data from an open-air parking lot at Shanghai Jiao Tong University, measuring 60m by 25m with 48 parking spaces. The area experiences frequent vehicle movement, along with pedestrians and non-motorized vehicles. A sensor suite, consisting of one repetitive scanning LiDAR, two non-repetitive LiDARs and two cameras, is installed on a building 19.25 meters above the site, providing full scene coverage (Fig. 2.b and c show the aerial view and sensor layout).

The repetitive scanning LiDAR is a Hesai 128-line LiDAR, providing dense and stable point clouds with a 360-degree rotation. It is tilted downward to capture approximately 160 degrees of data, with a capture rate of 1,538,000 points per second at 10 Hz. We also use line-downsampling to create virtual pseudo LiDARs with 64 and 32 lines for research purposes. Two non-repetitive scanning LiDARs, Livox Avia, each with a 70.4° × 77.2° field of view, are used to cover the entire site. Each LiDAR provides 240,000 laser points per second. Two cameras are also installed to monitor real-time operations at the site. Detailed sensor specifications are shown in Table 2, and samples of point clouds and images are in Fig 3. A high-precision GNSS and IMU system, with a localization error of about 1 cm at 100 Hz, is installed on the experimental vehicle to serve as the ground truth for vehicle localization. The vehicle is shown in Fig. 2.a.

TABLE 2. SENSOR SPECIFICATION

| Sensors | Details |
| --- | --- |
| 1 × Repetitive Scanning LiDAR | 160° ×40° FOV*, 128 channels, 200m range @ 10% reflectivity, 10Hz, 1,536,000* points/second (Hesai OT128) |
| 2 × Non-repetitive Scanning LiDAR | 70.4° × 77.2° FOV, 190m range @ 10% reflectivity, 10Hz, 24,000 points/second (Livox Avia) |
| 2 × Camera | 1280 × 720 resolution, 10Hz, auto exposure (Realtek Semiconductor Corp. USB camera) |
| 1 × GNSS&IMU | 100Hz, installed on the experimental vehicle |
| * Point rate calculated according to shrunk FOV. 1.538 M =160/360× 3.46M(nominal point rate) | |

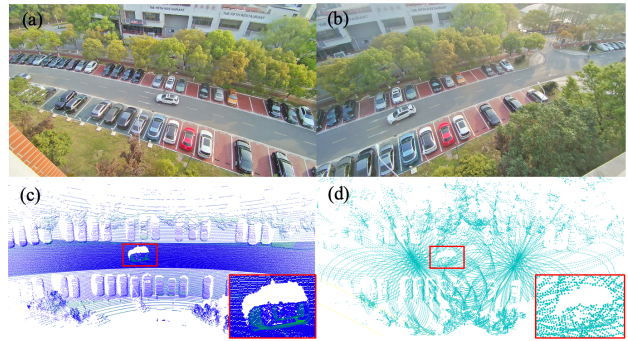

Fig. 3. Sample data of the dataset. (a) and (b) are images from two cameras. (c) and (d) are point clouds from 128-line repetitive scanning LiDAR and two non-repetitive scanning LiDAR Livox Avia respectively. The red box shows the point cloud falling on the experimental vehicle.

The data is organized in rosbag format, which facilitates processing with the ROS framework and allows users to leverage the relationship between consecutive frames for more accurate localization. Additionally, the format integrates seamlessly with ROS visualization tools (e.g., rviz), enabling easy visualization of data and localization results. Data from repetitive scanning LiDARs is saved in point cloud format, while data from non-repetitive scanning LiDARs uses the Livox CustomMsg format, which records per-point timestamps. This is crucial for correcting intra-frame distortions caused by dynamic objects moving at high speeds, which is used in Point-lio [9] and M-detector [11].

### B. Sensor Calibration

Firstly, time synchronization is performed as follows: both non-repetitive scanning LiDARs are triggered simultaneously by a single driver on the site-end computer at 10 Hz, ensuring synchronized timestamps. The 128-line LiDAR is also triggered at 10 Hz by the site-end computer, while the GNSS and IMU data are recorded by the vehicle-end computer at 100 Hz. The site-end and vehicle-end computers are connected to the same local area network and synchronized using the Network Time Protocol (NTP) [13]. Since the sampling frequency of the GNSS and IMU is significantly higher than that of the LiDAR, through time synchronization, we are able to obtain high-precision ground truth for vehicle localization corresponding to the timestamp of LiDAR data.

Next, coordinate system alignment is carried out by unifying the coordinate systems of the three LiDARs with the world coordinate system. The RANSAC method is used to estimate the ground plane, and the normal vector helps estimate the pitch, roll, and z coordinates of these LiDARs. The overlapping point clouds from different LiDARs are manually calibrated to align the x, y coordinates, and yaw angle of the LiDARs, thus aligning the two Livox LiDARs with the 128-line LiDAR. Through these steps, we align the extrinsic parameters of these three LiDARs into the same coordinate system.

## C. Ground Truth of Vehicle Localization

To obtain ground truth for vehicle localization, we first determine the transformation between the GNSS's UTM coordinate system and the site's world coordinate system. The process is as follows: First, we extract the vehicle's point cloud from each raw frame using a rule-based method (see Section IV) and data from the 128-line LiDAR, which provides dense and stable point clouds. Next, we match the extracted vehicle point cloud with a high-precision vehicle template, compute the bounding box, and identify the center of the bounding box as the vehicle center. This results in sequences of vehicle centers and GNSS-provided ground truths, which are then matched using timestamps. We optimize the UTM-to-world transformation by minimizing the distance error between matched points. Finally, the time- and coordinate-aligned GNSS results serve as the ground truth for vehicle localization. The vehicle's orientation angle, derived from high-precision IMU integration, is used as ground truth for orientation estimation.

## D. Data Statistics

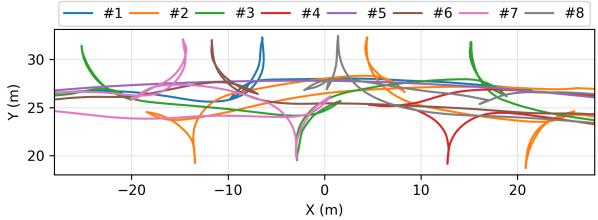

Fig. 4. Visualization of all vehicle trajectories in the dataset.

The dataset comprises 5,445 frames of point cloud data across 8 parking sequences, utilizing 11 distinct parking spots, as shown in Fig. 4. Vehicle orientation angles and speeds are analyzed and presented in Fig. 5. As the dataset primarily focuses on parking scenarios, with vehicles either driving or parking into spots, the orientation angles are predominantly aligned with the vertical and horizontal axes, while other directions exhibit a more even distribution. Regarding vehicle speeds, most data correspond to low-speed maneuvers typical of parking, with an average speed of 4.8 km/h.

Fig. 6 shows the statistical analysis of the number of point cloud points falling on the target vehicle per frame under different LiDAR configurations. Under the repetitive scanning LiDAR setup, a larger number of point clouds are observed on the vehicles, with the peak values around 1,100 points for the 128-line LiDAR and 800 points for the 64-line LiDAR. In contrast, with the non-repetitive scanning LiDAR configuration, the number of point clouds on vehicles is significantly lower, peaking at approximately 150 points.

## I. EXPERIMENTS AND DISCUSSIONS

### A. Baseline Experiments

Here, we choose four representative vehicle pose estimation methods as baseline methods and evaluate their performance on Bench-RNR Dataset. Below, we provide a detailed introduction to each baseline method.

To localize the target vehicle, we first segment the point cloud of the target vehicle in each frame. The process begins with scene modeling using a background modeling method, followed by background filtering to extract the foreground point cloud. Point cloud clustering is then applied to obtain instance-level point clouds. A greedy algorithm performs multi-object frame-to-frame association, enabling continuous tracking of both vehicles and pedestrians. This approach allows the point cloud of the target vehicle to be extracted from each frame of the original input point cloud. Finally, the segmented vehicle point clouds are fed into four typical vehicle pose estimation methods for vehicle localization respectively, with details as follows:

(1) The first method calculates the principal direction of the target vehicle's point cloud using Principal Component Analysis (PCA) [14], projects the point cloud onto this direction, and computes the oriented bounding box (OBB), referred to as "Seg+obb."

(2) The second method applies the convex-hull-based vehicle pose estimation from [15] to generate a bounding box for the segmented point cloud, referred to as "Seg + convex."

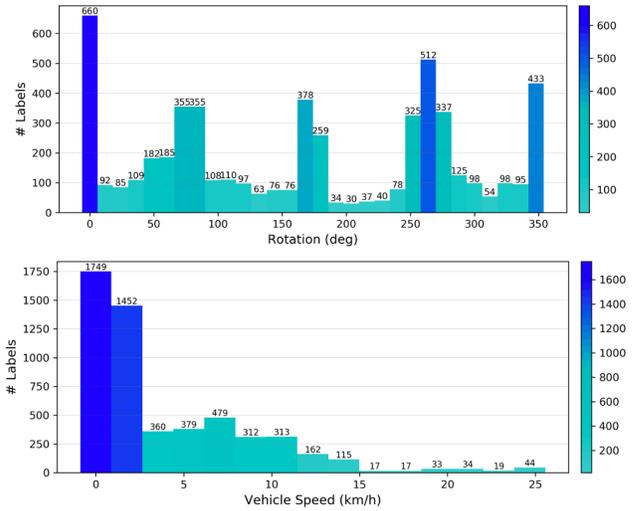

Fig. 5. Statistics analysis of Bench-RNR Dataset. The chart above shows the number of times vehicle's orientation falls into different intervals, while the chart below illustrates the distribution of vehicle speeds in the dataset.

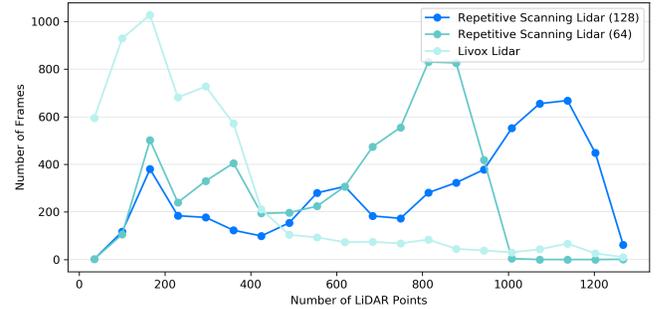

Fig. 6. Statistical analysis of the number of LiDAR points falling on the target vehicle under different LiDARs.

TABLE. 3. VEHICLE LOCALIZATION ACCURACY OF BASELINE METHODS

| Scanning Pattern | Method | Localization Error (m) | | | Heading Estimation Error (deg) | | |
|---|---|---|---|---|---|---|---|
| | | Mean | Std. Dev. | Max | Mean | Std. Dev. | Max |
| Non-Repetitive Scanning (2*Livox Avia) | Seg+obb | 16.67 | 7.80 | 44.63 | 3.69 | 2.52 | 10.87 |
| | Seg+convex | 15.56 | 7.68 | 41.29 | 1.35 | 1.32 | 7.54 |
| | PV-RCNN | 15.79 | 9.57 | 50.57 | 2.84 | 2.41 | 13.48 |
| | Register-Loc | **6.87** | **3.34** | **17.00** | **1.03** | **0.82** | **3.65** |
| Repetitive Scanning (OT-128) | Seg+obb | 14.23 | 7.32 | 38.20 | 4.77 | 3.13 | 11.15 |
| | Seg+convex | 13.06 | 6.80 | 36.44 | 1.60 | 1.30 | 6.13 |
| | PV-RCNN | 16.06 | 9.80 | 52.77 | 2.70 | 2.29 | 12.15 |
| | Register-Loc | **6.84** | **3.43** | **16.59** | **0.88** | **0.61** | **2.50** |

(3) The third method registers the segmented point cloud with a high-precision vehicle template and computes the bounding box of the template, referred to as "Register-Loc".
(4) The fourth method uses the PV-RCNN point cloud detection network [16]. Due to poor performance when applying the model trained on vehicle data directly to roadside data, we train the model with roadside point cloud data from the DAIR-V2X-I dataset [2] using the OpenPCDet framework [17], and perform inference on our roadside dataset.

We use the centers and orientations of the bounding boxes from these methods as the estimated center and orientation of the vehicle. Experiments are conducted on the four baseline methods using both repetitive and non-repetitive LiDAR data. Localization and heading estimation errors can be seen in Table 3. Below, we analyze and compare the performance of the four methods under the novel non-repetitive scanning.

The "Seg+obb" method has the largest localization and heading estimation errors among the four methods, with values of 16.67cm and 3.69 deg, respectively. This is mainly due to the sparse and uneven point cloud observations under non-repetitive scanning, leading to incomplete data that affects the bounding box size estimation and increases localization errors. Additionally, the uneven distribution of the point cloud causes PCA to estimate the principal direction with a bias, resulting in larger orientation angle errors.

The "Seg+convex" method ranks second in heading estimation performance, with an error of 1.35 deg, by effectively capturing the vehicle's L-shape feature. However, it still exhibits a significant localization error (15.56 cm), as the fitted cuboid often underestimates the actual vehicle length.

Although PV-RCNN, as a detector, has acquired prior knowledge about vehicle shapes through training, the localization error remains relatively large, at 15.79 cm. This may be due to the fact that PV-RCNN was trained on the DAIR-V2X-I dataset, which introduces a domain gap when directly applied to our dataset. It is worth noting that PV-RCNN has a relatively high rate of missed detection, with missing rates of 6.02% for repetitive scanning LiDAR and 12.25% for non-repetitive scanning LiDAR.

Register-Loc achieves the best localization and heading estimation results among these four baseline methods, with a localization error of 6.87 cm and a heading estimation error of 1.03 deg. This is because Register-Loc can fully utilize the 3D shape information of the vehicle in the point cloud template through point cloud registration, thereby achieving robust localization results even when the non-repetitive scanning point cloud observation is incomplete or sparse. However, this method requires the establishment of a high-precision vehicle point cloud template in advance, which entails additional equipment costs and computational overhead.

B. *Repetitive vs. Non-repetitive Scanning*

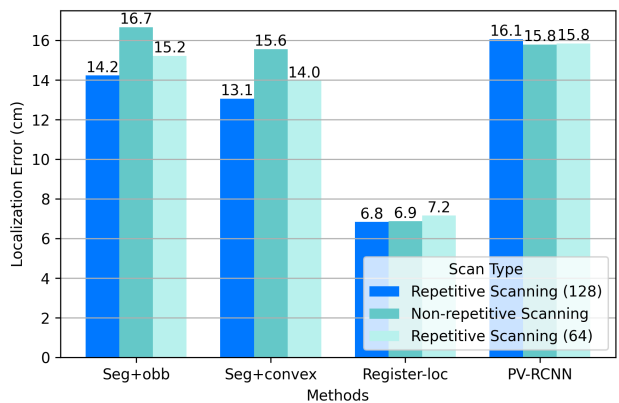

Fig. 7. The average localization errors of baseline methods across different LiDAR scanning patterns.

We also compare the localization performance of baseline methods using repetitive scanning LiDAR (128-line), repetitive scanning LiDAR (pseudo-64-line, downsampled from 128-line LiDAR) and non-repetitive scanning LiDAR (two Livox Avia). The results can be seen in Fig. 7.

It can be observed that the non-repetitive scanning LiDAR achieves results comparable to 128-line and 64-line repetitive scanning LiDAR across four different approaches, and even slightly outperforms the 64-line LiDAR in the "Register-Loc" and "PV-RCNN" methods. This indicates that non-repetitive scanning LiDAR, at a lower cost, can achieve localization performance comparable to repetitive LiDAR, providing centimeter-level accuracy. Under the "Register-Loc" method, the average localization error with non-repetitive scanning LiDAR is only 0.069 meters, which is sufficient for the safe operation of cloud-controlled vehicles. This highlights the potential of non-repetitive scanning in roadside vehicle localization applications.

However, it is important to recognize that the sparsity and uneven distribution of non-repetitive point clouds lead to fewer data points on vehicle targets and a less uniform distribution compared to repetitive scanning. Methods such as "Seg+obb" and "Seg+convex", which rely on the completeness of point cloud data, suffer from degraded localization performance when using non-repetitive scanning. Therefore, to effectively apply non-repetitive scanning LiDAR in roadside localization applications, it is recommended to use methods that incorporate prior knowledge of the target's complete shape, to mitigate the challenges posed by the inherent sparsity and unevenness of non-repetitive scanning.

## II. CONCLUSION

This paper addresses the current situation where non-repetitive scanning LiDAR has potential applications on the roadside but has not been widely adopted. We establish a real-vehicle dataset to benchmark the performance of repetitive and non-repetitive LiDAR in infrastructure-based vehicle localization. We construct a dataset containing 5,445 frames across eight trajectories, with both repetitive and non-repetitive LiDAR point clouds. We also develop four baseline methods to evaluate the localization performance of different LiDARs. The result shows that non-repetitive scanning LiDAR can achieve localization performance comparable to that of 128-line LiDAR and pseudo-64-line LiDAR, with a lower cost, sparser and unevenly distributed point clouds. This highlights the potential of non-repetitive scanning LiDAR for infrastructure-based vehicle localization and serves as a guideline for researchers configuring LiDAR for such applications.

For future work, we plan to collect data of non-repetitive scanning LiDAR from more roadside scenarios and annotate ground truth labels for not only vehicle localization but also object detection, further facilitating the widespread application of non-repetitive scanning LiDAR in roadside perception and localization.